\documentclass[conference]{IEEEtran}
\IEEEoverridecommandlockouts
\usepackage{cite}
\usepackage{amsmath,amssymb,amsfonts}
\usepackage{algorithmic}
\usepackage{graphicx}
\usepackage{textcomp}
\usepackage{xcolor}

\usepackage{hyperref}
\usepackage{amsmath}

\DeclareMathOperator*{\argmin}{arg\,min}

\def\BibTeX{{\rm B\kern-.05em{\sc i\kern-.025em b}\kern-.08em
    T\kern-.1667em\lower.7ex\hbox{E}\kern-.125emX}}
\begin{document}

\title{On the transferability of adversarial examples between convex and 01 loss models\\
}

\author{\IEEEauthorblockN{1\textsuperscript{st} Yunzhe Xue}
\IEEEauthorblockA{\textit{Department of Computer Science} \\
\textit{New Jersey Institute of Technology}\\
Newark, USA \\
yx277@njit.edu}
\and
\IEEEauthorblockN{2\textsuperscript{nd} Meiyan Xie}
\IEEEauthorblockA{\textit{Department of Computer Science} \\
\textit{New Jersey Institute of Technology}\\
Newark, USA\\
mx42@njit.edu}
\and
\IEEEauthorblockN{3\textsuperscript{rd} Usman Roshan}
\IEEEauthorblockA{\textit{Department of Computer Science} \\
\textit{New Jersey Institute of Technology}\\
Newark, USA\\
usman@njit.edu}
}

\maketitle

\begin{abstract}
The 01 loss gives different and more accurate boundaries than convex loss models in the presence of outliers. Could the difference of boundaries translate to adversarial examples that are non-transferable between 01 loss and convex models? We explore this empirically in this paper by studying transferability of adversarial examples between linear 01 loss and convex (hinge) loss models, and between dual layer neural networks with sign activation and 01 loss vs sigmoid activation and logistic loss. We first show that white box adversarial examples do not transfer effectively between convex and 01 loss and between 01 loss models compared to between convex models. As a result of this non-transferability we see that convex substitute model black box attacks are less effective on 01 loss than convex models. Interestingly we also see that 01 loss substitute model attacks are  ineffective on both convex and 01 loss models mostly likely due to the non-uniqueness of 01 loss models. We show intuitively by example how the presence of outliers can cause different decision boundaries between 01 and convex loss models which in turn produces adversaries that are non-transferable. Indeed we see on MNIST that adversaries transfer between 01 loss and convex models more easily than on CIFAR10 and ImageNet which are likely to contain outliers. We show intuitively by example how the non-continuity of 01 loss makes adversaries non-transferable in a dual layer neural network. We discretize CIFAR10 features to be more like MNIST and find that it does not improve transferability, thus suggesting that different boundaries due to outliers are more likely the cause of non-transferability. As a result of this non-transferability we show that our dual layer sign activation network with 01 loss can attain robustness on par with simple convolutional networks.
\end{abstract}

\begin{IEEEkeywords}
adversarial attacks, transferability of adversarial examples, 01 loss, stochastic coordinate descent, convolutional neural networks, deep learning
\end{IEEEkeywords}

\section{Introduction}
State of the art machine learning algorithms can achieve high accuracies in classification tasks but misclassify minor perturbations in the data known as as adversarial attacks 
\cite{goodfellow2014explaining,papernot2016limitations,kurakin2016adversarial,carlini2017towards,brendel2017decision}. Adversarial examples have been shown to transfer across models which makes it possible to perform transer-based (substitute model) black box attacks \cite{papernot2016transferability}. To counter adversarial attacks many defense methods been proposed with adversarial training being the most popular \cite{szegedy2013intriguing}. This is known to improve  robustness to adversarial examples but also tends to lower accuracy on clean test data that has no perturbations \cite{raghunathan2019adversarial,zhang2019theoretically}. Many previously proposed defenses have also shown to be vulnerable \cite{carlini2017towards,athalye2018obfuscated,ghiasi2020breaking} thus leaving adversarial robustness still an open problem in machine learning.

The 01 loss is known to be more robust to outliers than convex loss models \cite{xie2019,icml13optimize,bartlett04} and gives different boundaries in the presence of outliers. Could the difference in boundaries translate into non-transferability of adversarial examples between convex and 01 loss models? We study this in the setting of white box and substitute model black box attacks. 

Computationally 01 loss presents a considerable challenge because it is NP-hard to solve \cite{ben03}. Previous attempts \cite{nips01optimize,mixedint01,approx01,ijcnn01optimize,icml13optimize} lack on-par test accuracy with convex solvers and are slow and impractical for large multiclass image benchmarks. However, a recent stochastic coordinate descent method for linear 01 loss models \cite{xie2019} has shown to attain comparable accuracies to state of the art linear solvers like the support vector machine. We build upon this in our current work.

We propose a 01 loss dual layer neural network with sign activation which to the best of our knowledge is the first such network to be proposed. Since our network weights are real numbers our model is different from binarized neural networks whose weights are constrained to be +1 and -1 or 1 and 0 \cite{galloway2017attacking,courbariaux2016binarized,rastegari2016xnor}. We train our network with stochastic coordinate descent (along the lines of \cite{xie2019} and fully described in our Supplementary Material) and show that it achieves on-par test accuracy to equivalent convex models. We are now in a position to study transferability of adversarial examples between linear 01 loss and convex (hinge) loss models, and between dual layer neural networks with sign activation and 01 loss vs sigmoid activation and logistic loss. 

We proceed with white box attacks where the attacker has full knowledge of the model's parameters and  transfer based black box attacks via substitute models where the attacker has access only to label outputs of the model. We study transferability and black box attacks on image benchmarks MNIST \cite{lecun1998gradient}, CIFAR10 \cite{krizhevsky2009learning}, and Mini ImageNet (a ten class subset of the original ImageNet \cite{ILSVRC15}) where we make the following findings. 

\begin{itemize}
\item White box adversaries do not transfer effectively between convex and 01 loss on CIFAR10 and ImageNet datasets than they do on MNIST.
\item As a result of this, convex substitute model black box attacks are more effective on convex models than 01 loss ones on CIFAR10 and ImageNet.
\item 01 loss substitute model black box attacks are ineffective on all datasets and on all models due to their non-uniqueness.
\item Discretization of input features does not affect the non-transferability on CIFAR10, in fact it increases it and as a consequence makes 01 loss models even more robust to convex substitute model black box attacks than in the original feature space.
\item Our 01 loss models can attain comparable black box robustness to convolutional models than their convex counterparts.
\end{itemize}

\section{Methods}

\subsection{Background}
The problem of determining the hyperplane with minimum number of misclassifications
in a binary classification problem is known to be NP-hard \cite{ben03}.
In mainstream machine learning literature this is called minimizing the 01 loss
\cite{kernel01} given in Objective~\ref{obj1},

\begin{equation}
\frac{1}{2n}\argmin_{w,w_0} \sum_i (1-sign(y_i(w^Tx_i+w_0)))
\label{obj1}
\end{equation}

where $w \in R^d$, $w_0 \in R$ is our hyperplane, and $x_i \in R^d, y_i\in \{+1,-1\}.\forall i=0...n-1$ are our training data. Popular linear classifiers such as the linear support 
vector machine, perceptron, and logistic regression \cite{alpaydin} can be considered 
as convex approximations to this problem that yield fast gradient descent solutions \cite{bartlett04}. 
However, they are also more sensitive to outliers than the 01 loss \cite{bartlett04,icml13optimize,xie2019}
and more prone to mislabeled data than 01 loss \cite{manwani2013noise,ghosh2015making,lyu2019curriculum}. 

\subsection{A dual layer 01 loss neural network}
We extend the 01 loss to a simple two layer neural network with $k$ hidden nodes and sign activation that we call the MLP01 loss. This objective for binary classification can be given as

\begin{equation}
\small
\frac{1}{2n}\argmin_{W, W_0, w,w_0} \sum_i (1-sign(y_i(w^T(sign(W^Tx_i+W_0))+w_0)))
\label{obj2}
\end{equation}

where $W \in R^{d\times k}$, $W_0 \in R^k$ are the hidden layer parameters, $w\in R^k, w_0\in R$ are the final layer node parameters, $x_i \in R^d, y_i\in \{+1,-1\}.\forall i=0...n-1$ are our training data, and $sign(v\in R^k)=(sign(v_0), sign(v_1),...,sign(v_{k-1}))$. While this is a straightforward model to define optimizing it is a different story altogether. Optimizing even a single node is NP-hard which makes optimizing this network much harder. Note that our weights are real numbers as opposed to binarized neural networks whose weights are constrained to be +1 and -1 or 1 and 0 \cite{galloway2017attacking,courbariaux2016binarized,rastegari2016xnor}. 



\subsection{Stochastic coordinate descent for 01 loss}
We solve both problems with stochastic coordinate descent based upon earlier work \cite{xie2019}. We initialize all parameters to random values from the Normal distribution with mean 0 and variance 1. We then randomly select a subset of the training data (known as a batch) and perform the coordinate descent analog of a single step gradient update in stochastic gradient descent \cite{bottou2010large}. We first describe this for a linear 01 loss classifier which we obtain if we set the number of hidden nodes to zero. In this case the parameters to optimize are the final weight vector $w$ and the threshold $w_0$. 

When the gradient is known we step in its negative direction by a factor of the learning rate: $w=w-\eta\nabla(f)$ where $f$ is the objective. In our case since the gradient does not exist we randomly select $k$ features (set to 64, 128, and 256 for MNIST, CIFAR10, and ImageNet in our experiments), modify the corresponding entries in $w$ by the learning rate (set to 0.17) one at a time, and accept the modification that gives the largest decrease in the objective. Key to our search is a heuristic to determine the optimal threshold each time we modify an entry of $w$. In this heuristic we perform a linear search on a subset of the projection $w^Tx_i$ and select $w_0$ that minimizes the objective.

\begin{figure}[h]
  \centering
  \includegraphics[trim=80 50 0 70, clip, scale=.27]{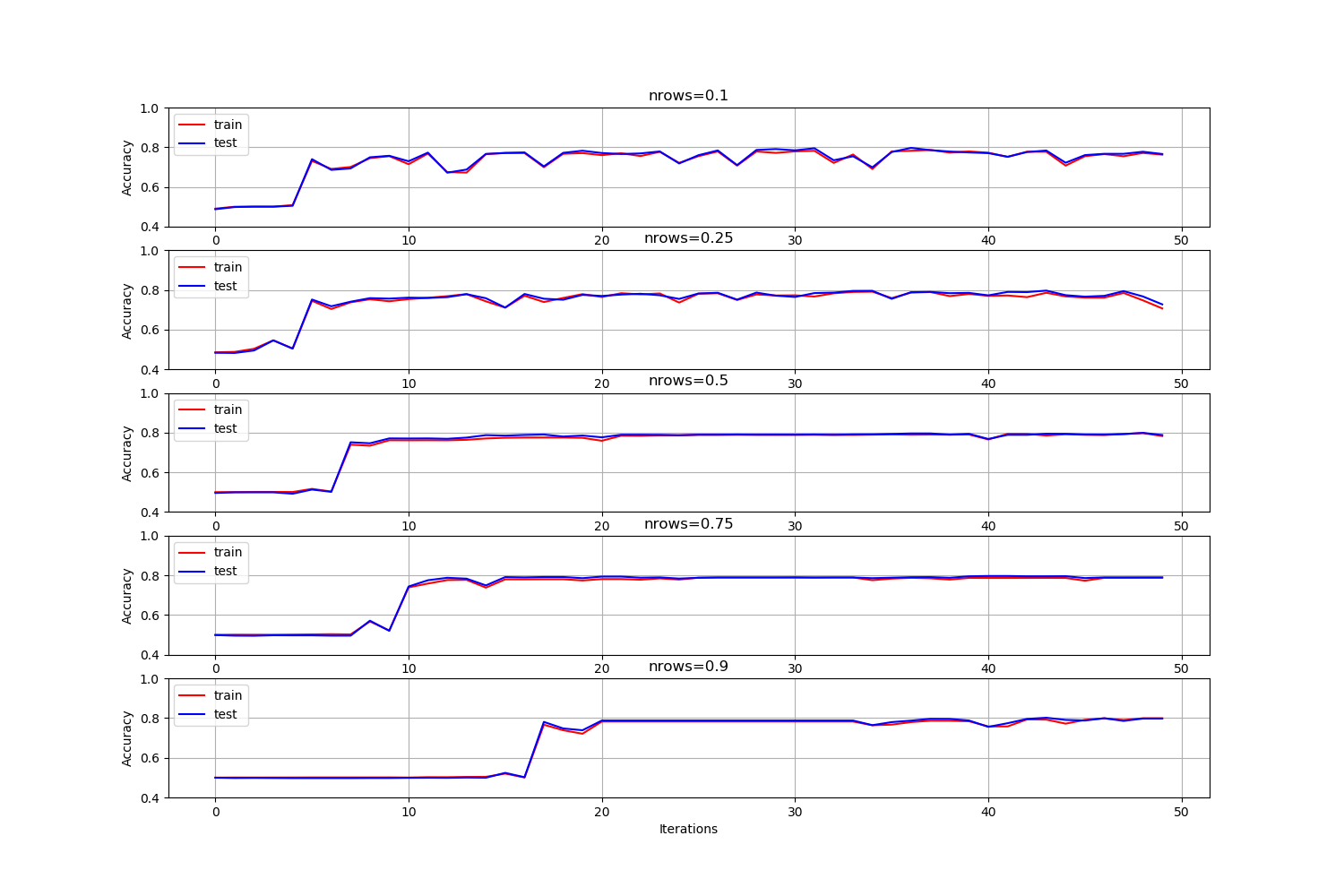} 
  \caption{Train and test accuracy of our stochastic coordinate descent on CIFAR10 class 0 vs 1 with different batch sizes (denoted as nrows). \label{nrows}}
\end{figure}

We repeat the above update step on randomly selected batches for a specified number of iterations given by the user. In Figure~\ref{nrows} we show the effect of the batch size (as a percentage of each class to ensure fair sampling) on a linear 01 loss search on CIFAR10 between classes 0 and 1. We see that a batch size of 75\% reaches a train accuracy of 80\% faster than the other batch sizes. Thus we use this batch size in all our experiments going forward.

We also see that for this batch size the search flattens after 15 iterations (or epochs as given in the figure). We run 1000 iterations to ensure a deep search with an intent to maximize test accuracy. For imbalanced data (that appears in the one-vs-all design) we find that optimizing a balanced version of our objective for half the iterations followed by the default (imbalanced) version gives a lower objective in the end.

In a dual layer network we have to optimize our hidden nodes as well. In each of the 1000 iterations of our search we apply the same coordinate update described above, first to the final output node and then a randomly selected hidden node. In preliminary experiments we find this to be fast and almost as effective as optimizing all hidden nodes and the final node in each iteration. 

Our intuition is that by searching on just the sampled data we avoid local minima and across several iterations we can explore a broad portion of the search space. Throughout iterations we keep track of the best parameters that minimize our objective on the full dataset. Below we provide full details of our algorithms. 

The problem with our search described above is that it will return different solutions depending upon the initial starting point. To make it more stable we run it 32 times from different random seeds and use the majority vote for prediction. 

We extend both our linear and non-linear models to a simple one-vs-all approach for multiclass classification. For a dataset with $k$ classes we create $32$ one-vs-all classifiers for each of the $k$ classes. From the 32 models we can obtain frequency outputs for a test point using simple counting and use them as confidence scores for each class $k$. From this we output the predicted class as the one with the highest confidence. This is similar in spirit to the typical convex softmax objective used in convex neural networks except that there we can optimize to obtain the exact confidences given by sigmoid probabilities.

\subsection{Implementation, experimental platform, and image data}
We implement our 01 loss models in Python and Pytorch \cite{pytorch}, and both MLP and SVM (LinearSVC class) in scikit-learn \cite{scikit}. We optimize MLP with stochastic gradient descent that has a batch size of 200, momentum of 0.9, and learning rate of 0.01 (.001 for ImageNet data). We ran all experiments on Intel Xeon 6142 2.6GHz CPUs and NVIDIA Titan RTX GPU machines (for parallelizing multiple votes). Our SCD01 and MLP01 source codes, supplementary programs, and data are available from \url{https://github.com/zero-one-loss/mlp01}. 


We experiment on three popular image benchmarks: MNIST \cite{lecun1998gradient}, CIFAR10 \cite{krizhevsky2009learning}, and ImageNet \cite{ILSVRC15}. Briefly MNIST is composed of grayscale handwritten digits each of size $28\times28$ with 60000 training images and 10000 test and CIFAR10 has $32\times32$ color images with 50000 training and 10000 test. ImageNet is a large benchmark with 1000 classes and color images of size $256\times 256$. We extract images from10 random classes and split them to give a training set of 6144 images and test set of 6369. We normalize each image in each benchmark by dividing each pixel value by 255.

\section{Results}
We refer to our linear (no hidden layer) and non-linear (single hidden layer with 20 nodes) models as SCD01 and MLP01 respectively. 
As convex counterparts we select the linear support vector machine (with a cross-validated regularization parameter) denoted as SVM and a dual layer 20 hidden node neural network with logistic loss (MLP). For multiclass we use one-vs-all for all four methods. We use the majority vote of 32 runs for our 01 loss models to improve stability and do the same for SVM and MLP by majority voting on 32 bootstrapped samples. 
Our SCD01 and MLP01 source codes, supplementary programs, and data are available from \url{https://github.com/zero-one-loss/mlp01}. 

We refer to the accuracy on the test data as clean data test accuracy. An incorrectly classified adversarial example is considered a successful attack whereas a correctly classified adversarial is a failed one. Thus when we refer to accuracy of adversarial examples it is the same as $100-attack success rate$. The lower the accuracy the more effective the attack.

\subsection{Clean accuracy and runtimes}
Before going into robustness we first compare the clean test data accuracies and training runtimes of our 01 loss models to their convex counterparts. In Table~\ref{cleanacc} we see that ensembling SVM and MLP models does not improve the test accuracy over single runs, thus we use a shared weight MLP network with 400 nodes on ImageNet to boost accuracy there. 
In fact the SVM boundary depends only upon the support vectors and so each ensemble will be the same as long as the support vectors are included. As a reminder we ensemble by taking the majority vote on multiple bootstrapped samples.

The 01 loss models improve considerably in all three datasets by ensembling. This is not too surprising since 01 loss is non-unique and will give different solutions when ran multiple times from different initializations. As a result of ensembling their accuracy is comparable to their convex peers. This makes it easier to compare their robustness since we don't have to worry about the robustness vs accuracy tradeoff \cite{raghunathan2019adversarial,zhang2019theoretically,tsipras2018robustness}. 

\begin{table}[!h]
  \caption{Accuracy of  01 and convex counterparts on clean test data \label{cleanacc}}
  \centering
  \begin{tabular}{lllll} \hline
      &  \multicolumn{4}{c}{Single run}\\  
               & SVM  & SCD01 & MLP & MLP01  \\
    MNIST &  91.7  &	83.7 & 97.6 & 91.2\\
    CIFAR10 &  39.9 & 30.7 & 50.2 & 34.3 \\
    Mini ImageNet  &  26 &	25 &  32 & 25.5  \\ \hline
     & \multicolumn{4}{c}{32 votes} \\ 
     & SVM   & SCD01  & MLP  & MLP01 \\
MNIST  & 91.7 &  90.8 & 97.1 & 96 \\
CIFAR10  & 40.2 & 39.7 & 47.4 & 46.4 \\  
    &  MLP400 & SCD01  & & MLP01 \\
     &   single run  & &  & \\
 Mini ImageNet    & 36 &  34.7 & & 41\\ \hline

  \end{tabular}
\end{table}

In Table~\ref{runtimes} we show the runtime of a single run of our 01 loss and convex models on class 0 vs all for each of the three datasets. We don't claim the most optimized implementation but our runtimes are still somewhat comparable to the convex loss models. Interestingly the convex models take much longer on complex and higher dimensional images in ImageNet compared to MNIST. Our 01 loss model runtimes are similar on MNIST and CIFAR10 because their sizes are similar. On Mini ImageNet since it has fewer training samples than MNIST and CIFAR10 the 01 loss runtimes are also lower.

\begin{table}[!h]
  \caption{Runtimes in seconds of single runs on class 0 vs all. Mini ImageNet runtimes for are MLP with 400 hidden nodes. \label{runtimes}}
  \centering
  \begin{tabular}{llllll} \hline
               & SVM  & SCD01 & MLP & MLP01  &  \\
    MNIST &  0.8  &	171 & 64 & 875  \\
    CIFAR10 &  80 & 150 & 267 & 838  \\     
    Mini ImageNet  & 659 &	 83 & 8564 & 199 \\ \hline

  \end{tabular}
\end{table}

\subsection{White box attacks}
In this section we study white box attacks just for binary classification on classes 0 and 1 in each of the three datasets. We use single runs of each of the four models to generate adversaries using the model parameters. We use the same white box attack method \cite{papernot2016transferability} for SVM and SCD01 since both are linear classifiers: for a given datapoint $x$ and its label $y$ the adversary is $x' = x + \epsilon(-y)sign(w)$ where $sign(w)=(sign(w_1),sign(w_2),...,sign(w_{d}))$ and  $\epsilon$ is the distortion. 

For MLP we use the fast gradient sign method (FGSM) \cite{goodfellow2014explaining}. In this method we generate an adversary using the sign of the model gradient $x_{adv}=x+\epsilon sign(\nabla_x f(x,y)$ where $f$ is the model objective and $\nabla_x f(x,y)$ is the model gradient with respect to the data $x$. We generate white box adversaries for MLP01 with a simple heuristic: for each hidden node $w_k$ we modify the input as $x' = x + \epsilon(-y')sign(w_k)$ (where $y'=sign(w_k^Tx)$ is the output of $x$ from the hidden node $w_k$) and accept the first modification that misclassifies $x$ in the final node output. If $x$ is already misclassified or if none of the hidden node distortions misclassify it we distort with a randomly selected hidden node. We provide the full algorithm in the Supplementary Material. We use $\epsilon$ values on MNIST, CIFAR10, and ImageNet that are typical in the literature. 
 
In Table~\ref{whitebox} we see that the clean accuracies of our 01 loss models are comparable to the convex counterparts. As expected adversaries from the source on the same target are effective except for MLP01. More interestingly, while adversaries from SVM and MLP affect each other considerably they are far less pronounced on SCD01 and MLP01. We see this very clearly on CIFAR10 where both SVM and MLP adversaries have almost 0\% accuracy when attacking each other indicating high transferability \cite{papernot2016transferability}. But SVM and MLP adversaries on SCD01 and MLP01 have a far less effect in this dataset. Adversaries from MLP attain a 63.7\% accuracy on MLP01 and 43.5\% on SCD01. Another interesting observation is that adversaries barely transfer between SCD01 and MLP01. We see similar behavior on Mini ImageNet and to a lesser degree on MNIST.

\begin{table}[!h]
  \caption{Accuracy of adversaries made by source in the first column targeting models in the top row  (class 0 vs 1). \label{whitebox}}
  \centering
  \begin{tabular}{lllll} \hline
               & SVM  & SCD01 & MLP & MLP01 \\ \hline
           & \multicolumn{4}{c}{MNIST $\epsilon=.3$}   \\
    Clean &  100 &	99.9 &	100 & 100 \\
    SVM & 11.9 &	8.1	&40.4	&43.5  \\ 
    SCD01  &  97 &	0	& 98.5 &  53.2   \\
    MLP   & 25.5	 & 16.1 &	31 & 	42.3  \\
    MLP01   &  99.9	& 99.8 &	99.6 &	69.5  \\ \hline
       & \multicolumn{4}{c}{CIFAR10 $\epsilon=.0625$} \\
   Clean &  82.2 &	81.1	& 88.7 &	84.2 \\
    SVM   &  0	& 41.3&	0.5 & 	70.1	\\
    SCD01 &   76 & 	0.8	& 86 &	84.5 \\
    MLP  &  0	& 43.5 &	0.4 & 	63.7 \\
    MLP01 &  81.7	& 80 & 	88.5 &	66.9 \\ \hline
          &        \multicolumn{4}{c}{Mini ImageNet $\epsilon=.0625$}   \\
    Clean &  60.7 &	67.5 & 	66.1 &	68.7 \\
    SVM & 0	& 54.9 &	21.2 &	53.8 \\ 
    SCD01     &   58.6 &	1	& 65	& 60.3 \\
    MLP   & 0.5	& 42	& 21.6 &	52.3 \\
    MLP01   &  60.8 & 	65.1 & 	65.8 &	35.7 \\ \hline
  \end{tabular}
\end{table}

\begin{figure}[h]
  \begin{tabular}{l}
  \includegraphics[trim=125 180 200 80,clip,scale=.4]{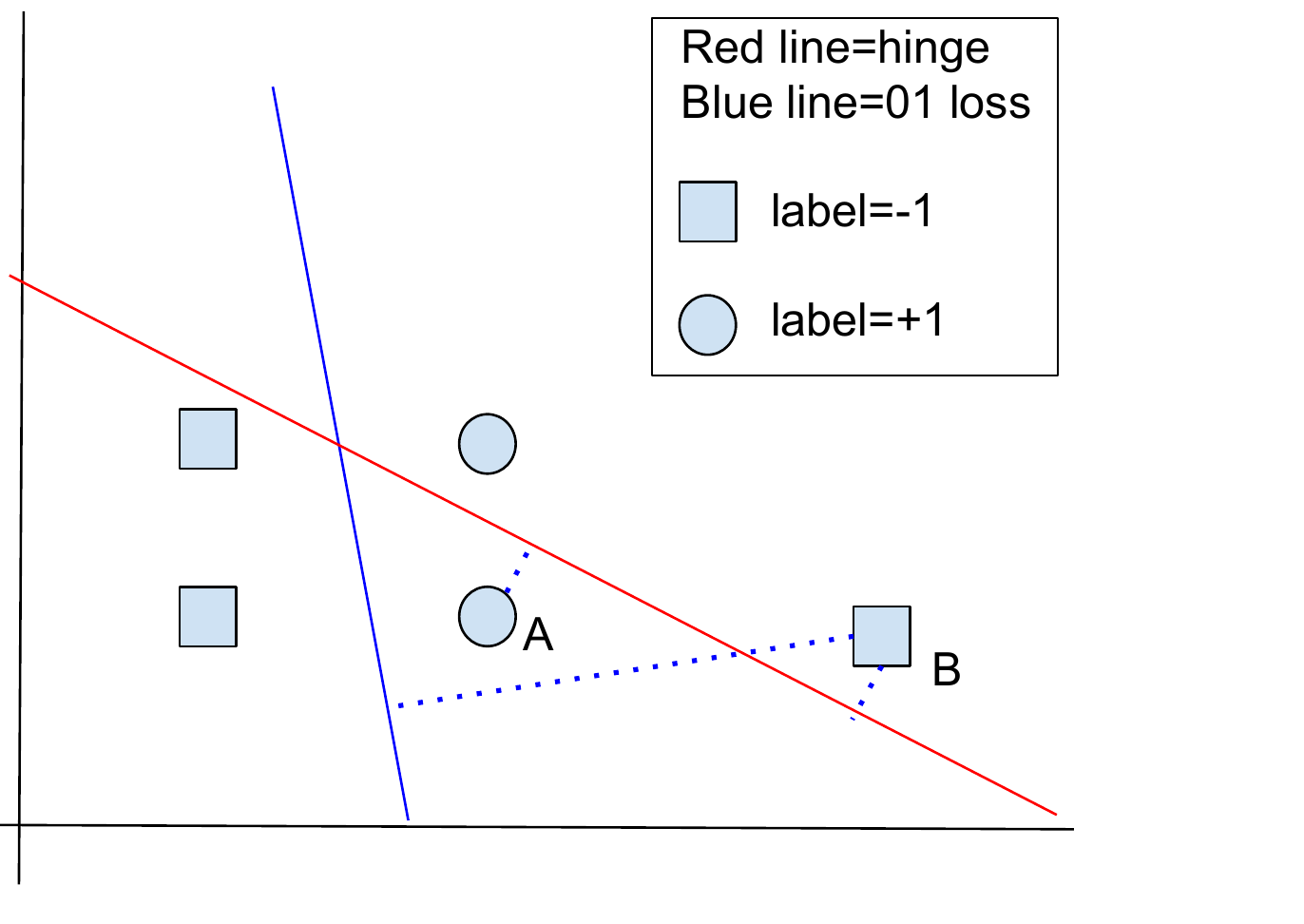} \\
  \footnotesize
   (a) Point $B$ is misclassified by the blue line but its hinge loss $max(0,1-yw^Tx)$\\
  \footnotesize   in dotted lines is higher than the combined loss of 
  points $A$ and $B$ both\\
  \footnotesize  
  \footnotesize misclassified by the red boundary. Thus hinge favors the red skewed line.\\

  \includegraphics[trim=70 220 100 10,clip,scale=.4]{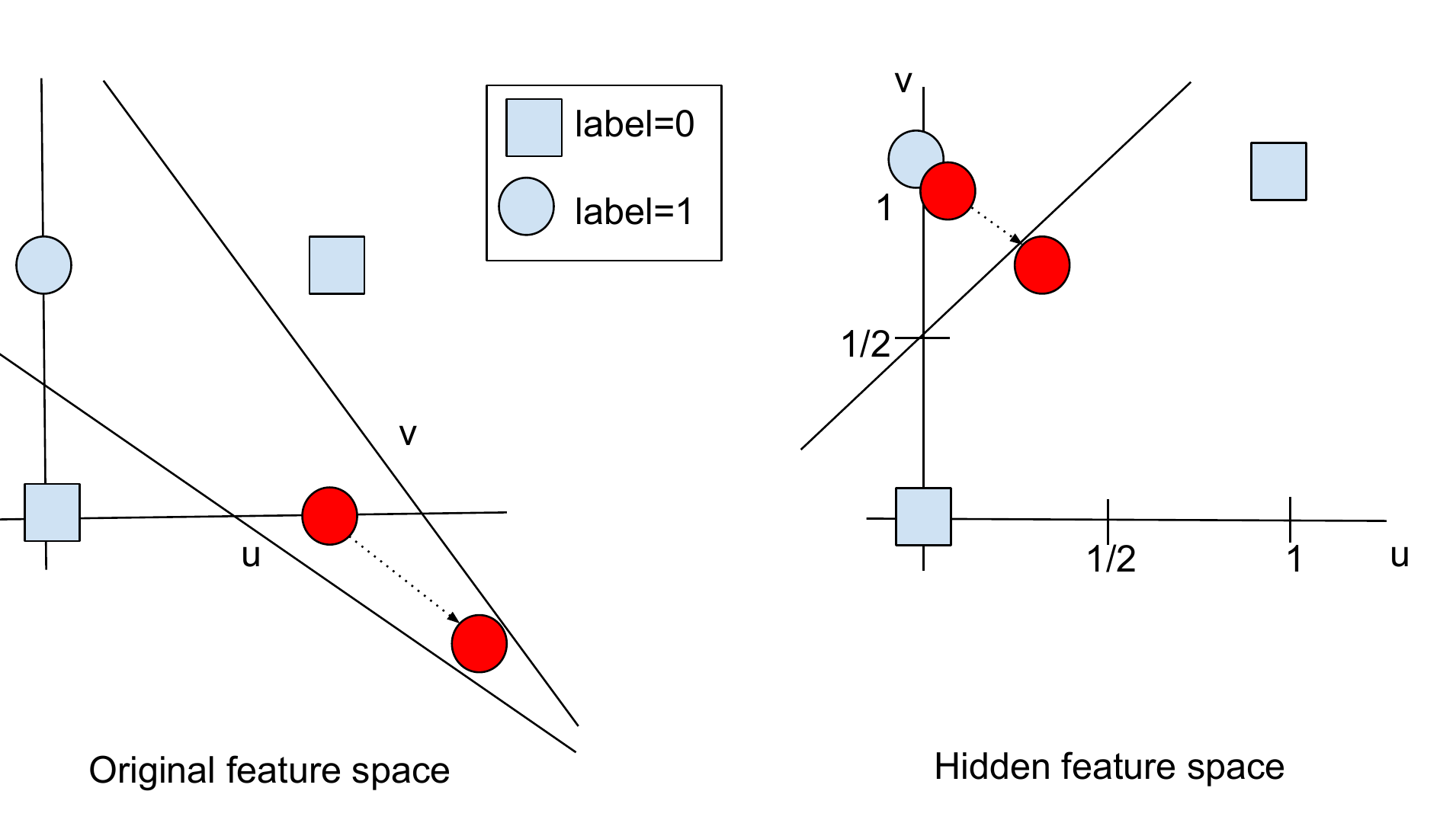} \\
  \footnotesize
 (b) The logistic activation $\frac{1}{1+e^{-w^Tx}}$ in the original space gives a linear\\
\footnotesize  separation in the hidden space. If the red circle moves towards the "corner"\\
\footnotesize  of the boundaries its distance to $u$ and $v$ decreases. This in turn makes its\\
\footnotesize   activation values approach $1/2$ and it is misclassified in the hidden space.\\
\footnotesize  In 01 loss activation 
\footnotesize the red circle does not get  affected in the hidden space.
 \end{tabular}
  \caption{Toy example showing different 01 loss and hinge boundaries, and adversarial examples in simple logistic loss network \label{white2}}
\end{figure}

We argue that the difference of loss functions (01 vs convex) may be responsible for different boundaries and non-transferability. We illustrate this in two examples. First we see the effect of outliers on 01 loss and hinge loss linear classifiers. Recall that the hinge loss is $max(0,1-yw^Tx)$ where $y$ is the label and $w^Tx$ is the prediction of $x$ given by the classifier $w$. In Figure~\ref{white2}(a) the misclassified outlier forces the hinge loss to give a skewed linear boundary with two misclassifications. This happens because even though the two points are misclassified by the red boundary they are closer to it than the single misclassified one is to the blue one. The 01 loss is unaffected by distances and thus gives the blue boundary with one misclassification. Since the two boundaries have different orientations their adversaries are also likely to be different. In a dataset like MNIST where our accuracies are high we don't expect many misclassified outliers and thus boundaries are unlikely to be different. As a result we see that many adversaries transfer between SVM and SCD01 on MNIST. But on CIFAR10 and Mini ImageNet, which are more complex and likely to contain misclassified outliers, we expect different boundaries which in turn gives fewer adversaries that transfer between the two. 

Next we see the difference of convex and 01 loss in simple two hidden node network. In Figure~\ref{white2}(b) we see two hyperplanes $u$ and $v$ on the left whose logistic outputs give the hidden feature space on the right. The two hyperplanes $u$ and $v$ represent two hidden nodes in a two layer network. Recall that the logistic activation $\frac{1}{1+e^{-w^Tx}}$ (where $w^Tx$ is prediction of $x$ given by $w$) is similar to 01 loss: for large values of $|w^Tx|$ it approaches 0 or 1 depending upon the sign of $w^Tx$ and approaches $\frac{1}{2}$ as $|w^Tx|$ approaches 0. Thus if we move the red circle towards the "corner" in the original feature space (as shown in Figure~\ref{white2}(b)) its outputs from $u$ and $v$ approach $\frac{1}{2}$ in the hidden space. Consequently it crosses the linear boundary in the hidden space and becomes adversarial. However if the activation is 01 loss the red point remains unmoved in the hidden space. In fact in 01 loss a datapoint's value in the hidden space changes only if we cross a boundary in the original space.

While both examples are not formal proofs they give some intuition of why fewer adversarial examples  transfer between 01 loss and convex loss compared to between just convex. In particular we see that for 01 loss a datapoint becomes potentially adversarial if and only if it crosses a boundary in the original feature space whereas this is not true for convex losses.

Interestingly we also see MLP01 adversaries don't transfer to the other three models. When applied to MLP01 the adversaries lower its accuracy relative to clean data but to lesser degree than other models attacking themselves. Thus our white box attack method for MLP01 may not be the most powerful one leaving this an open problem.

\subsection{Substitute model black box attacks}
We see that white box adversaries don't transfer between convex and 01 loss but can we attack a 01 loss model with a convex substitute model \cite{papernot2016transferability}? In this subsection we consider binary and multiclass classification on all three datasets. For all four methods we use 32 votes and one-vs-all multiclass classification. We use adversarial data augmentation \cite{papernot2016transferability} to iteratively train a substitute model trained on label outputs from the target model. In each epoch we generate white box adversaries targeting the substitute model with the FGSM method \cite{goodfellow2014explaining} and evaluate them on the target. Note that our black box attack is untargeted, we are mainly interested in misclassifying the data and not the misclassification label. See  Supplementary Material for the full substitute model learning algorithms but it is essentially the method of Papernot et. al. \cite{papernot2016transferability}. 

\subsubsection{Convex substitute model}

\begin{figure}[!h]
  \centering
  \begin{tabular}{c}
  \includegraphics[scale=.33]{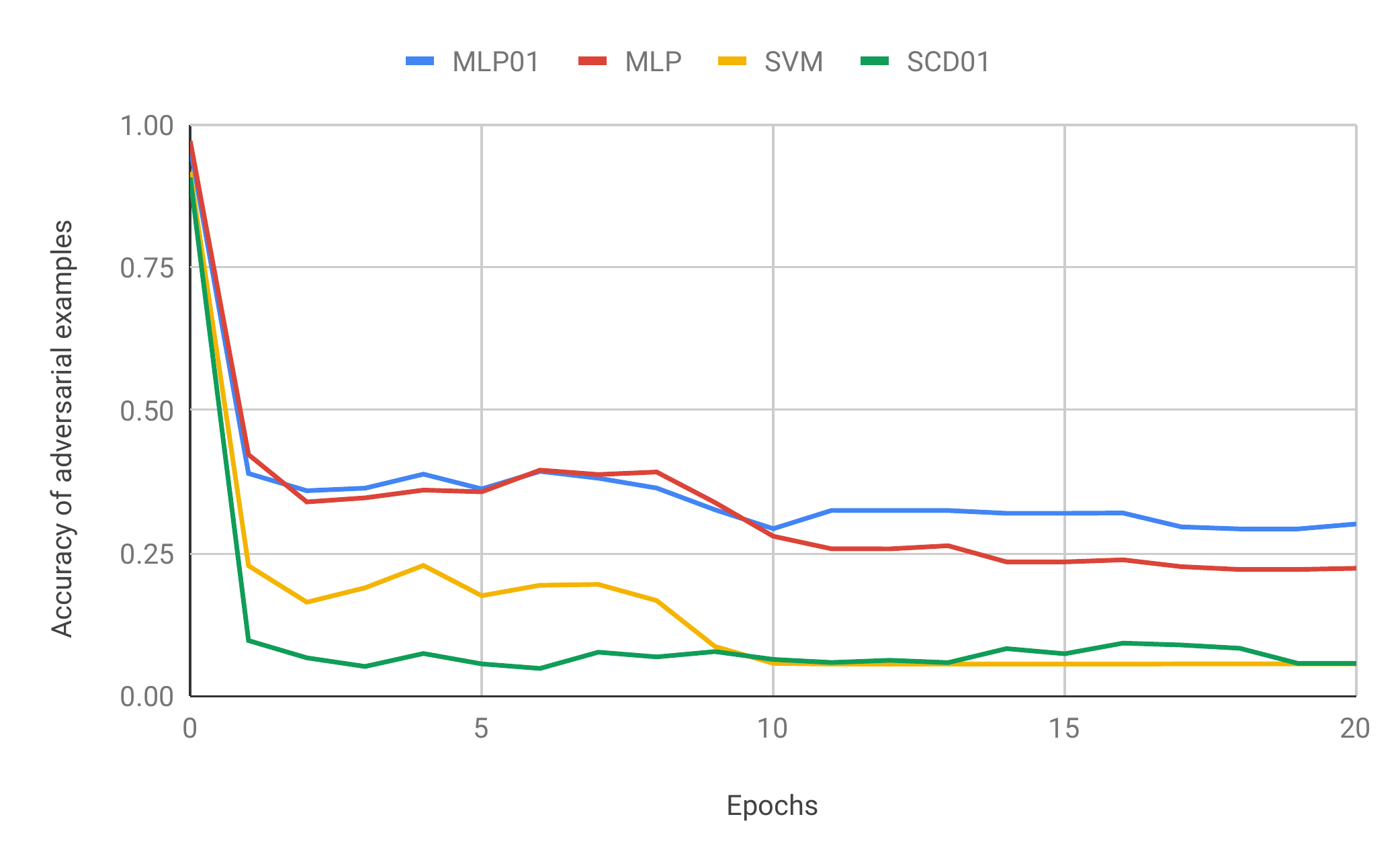} \\
  \footnotesize (a) MNIST $\epsilon=.2$ \\
  \includegraphics[scale=.33]{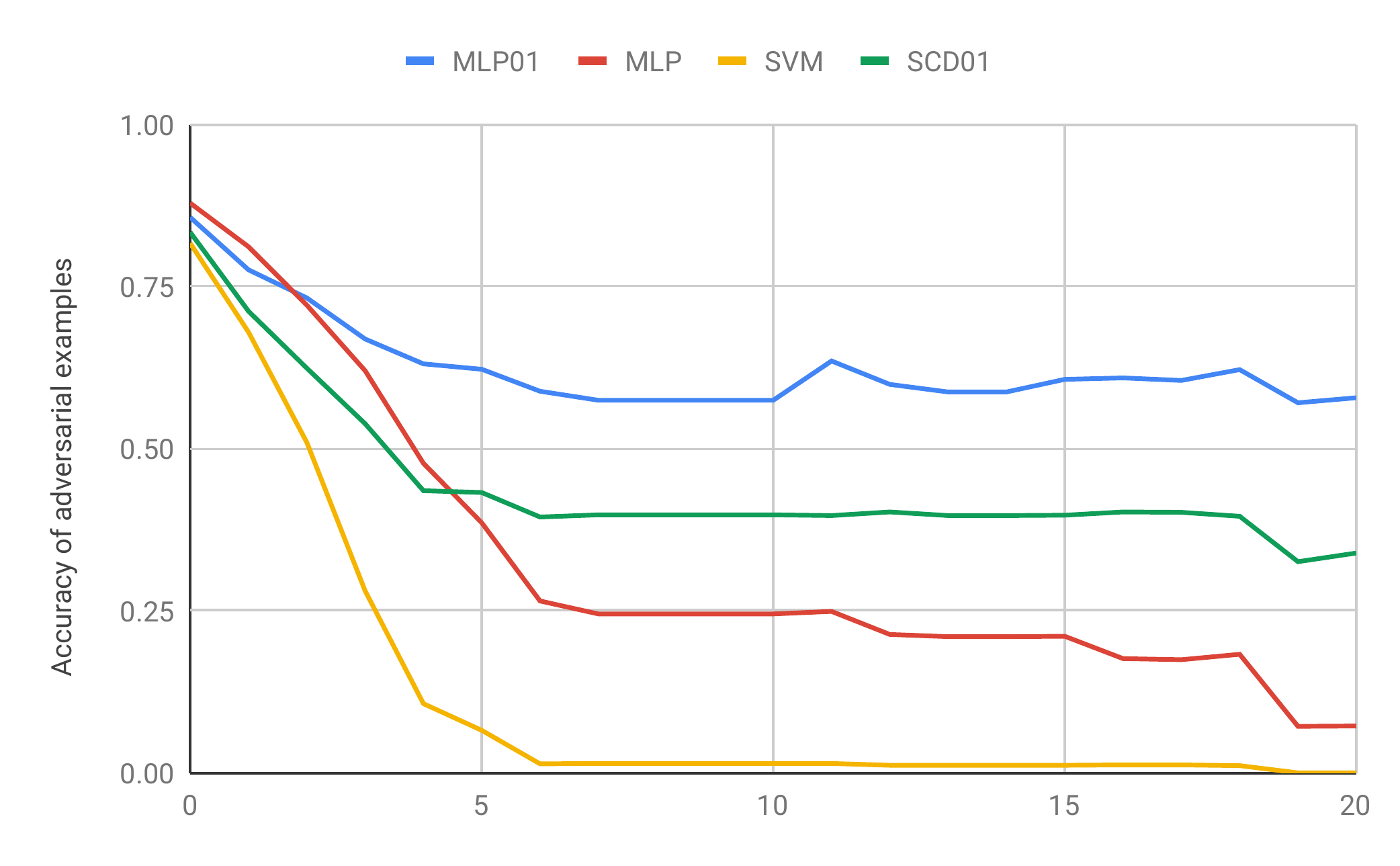} \\
   \footnotesize (b) CIFAR10 binary (class 0 and 1) $\epsilon=.0625$ \\
     \includegraphics[scale=.33]{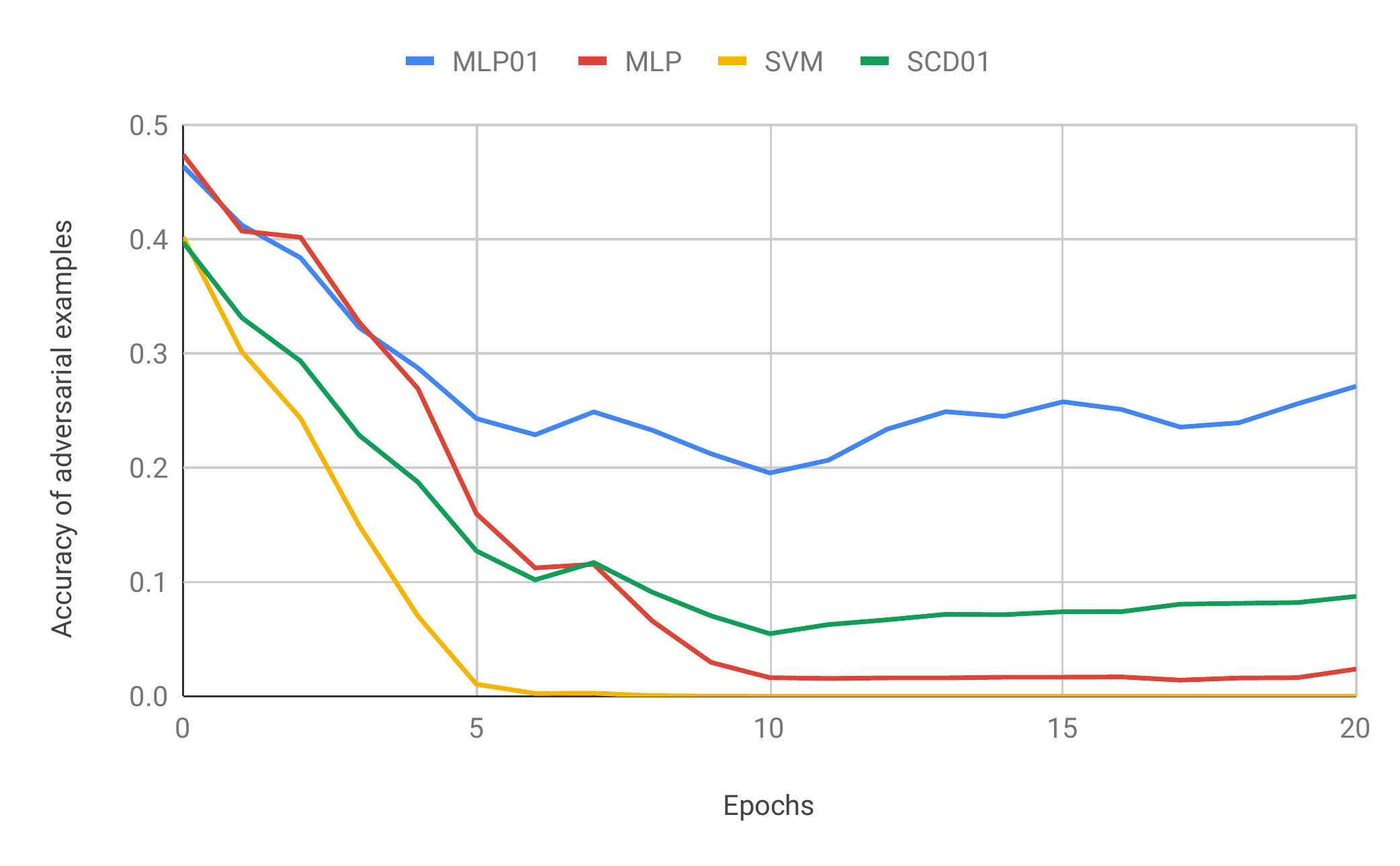} \\
   \footnotesize (c) CIFAR10 $\epsilon=.0625$ \\
   \includegraphics[scale=.33]{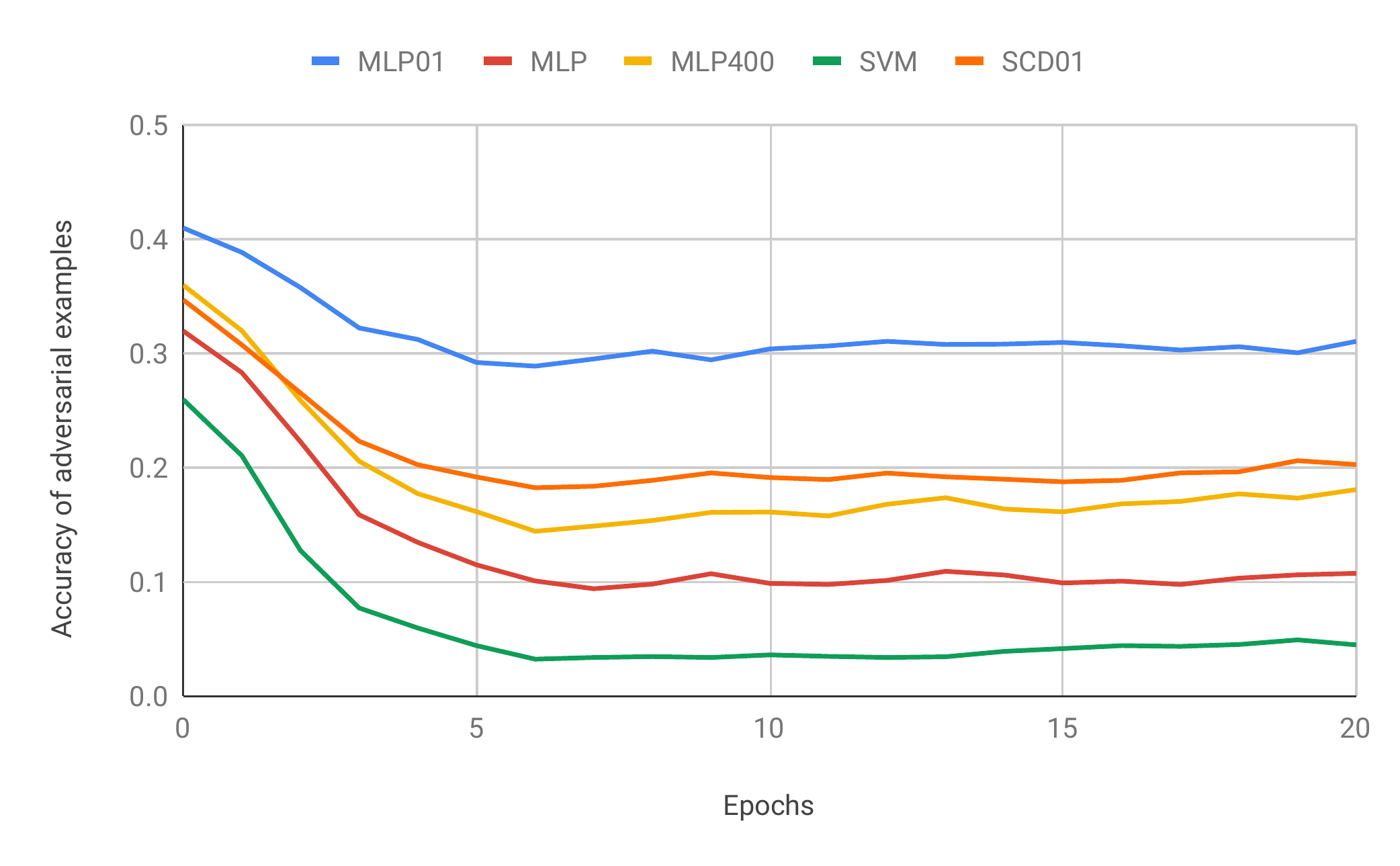} \\
  \footnotesize (d) Mini ImageNet $\epsilon=0.0625$ \\
 \end{tabular}
  \caption{Multiclass untargeted black box attack with a dual 200 node hidden layer logistic loss network as the substitute model. In epoch 0 are the clean test accuracies. \label{blackbox}}
\end{figure}

In Figure~\ref{blackbox} we see the accuracy of target models on adversaries generated from a convex substitute model. Specifically we use a dual hidden layer neural network with logistic loss and 200 nodes in each hidden layer as the substitute model. Like in the white box attacks we use $\epsilon$ values commonly used on these datasets. In MNIST (Figure~\ref{blackbox}(a)) we see a rapid drop in accuracy in the first few epochs and somewhat flat after epoch 10. We don't see a considerable difference between the 01 loss and convex sibling models on MNIST although MLP01 has the highest accuracy.

On CIFAR10 and Mini ImageNet we see much more pronounced differences. In CIFAR10 binary classification (Figure~\ref{blackbox}(b)) we see that even though both MLP and MLP01 start off with  clean test accuracies of 88\% and 86\% respectively, at the end of the 20th epoch MLP01 has 58\% accuracy on adversarial examples while MLP has 7\% accuracy. We see similar results on Mini ImageNet binary classification in the Supplementary Material. In CIFAR10 multiclass (Figure~\ref{blackbox}(c)) at the end of the 20th epoch the difference in accuracy between MLP and MLP01 is 24\% even though both methods start off with about the same accuracy on clean test data. Similarly on Mini ImageNet MLP01 is 20.7\% higher in accuracy than MLP in the 20th epoch. This is particularly interesting since MLP01 started off with a higher accuracy on Mini ImageNet and in general we expect more accurate models to be less robust \cite{raghunathan2019adversarial,zhang2019theoretically,tsipras2018robustness}. However that is not the case here. Even if we give MLP the advantage of 400 hidden nodes in a shared weight network instead of one-vs-all, its accuracy in the 20th epoch is 13\% lower than MLP01.

We have already seen earlier in white box attacks that adversaries transfer between SVM and SCD01 on MNIST but not so much on CIFAR10 and Mini ImageNet. The same phenomena can be used to explain the results we see here. On MNIST the convex substitute model can attack SCD01 and MLP01 as effectively as convex models due to better transferability on MNIST. Due to poor transferability on CIFAR10 and Mini ImageNet we see that the attack is less effective on SCD01 and MLP01. In the next subsection we explore what happens if the substitute model is SCD01.

\subsubsection{01 loss substitute model}
\begin{figure}[!h]
  \centering
  \includegraphics[scale=.33]{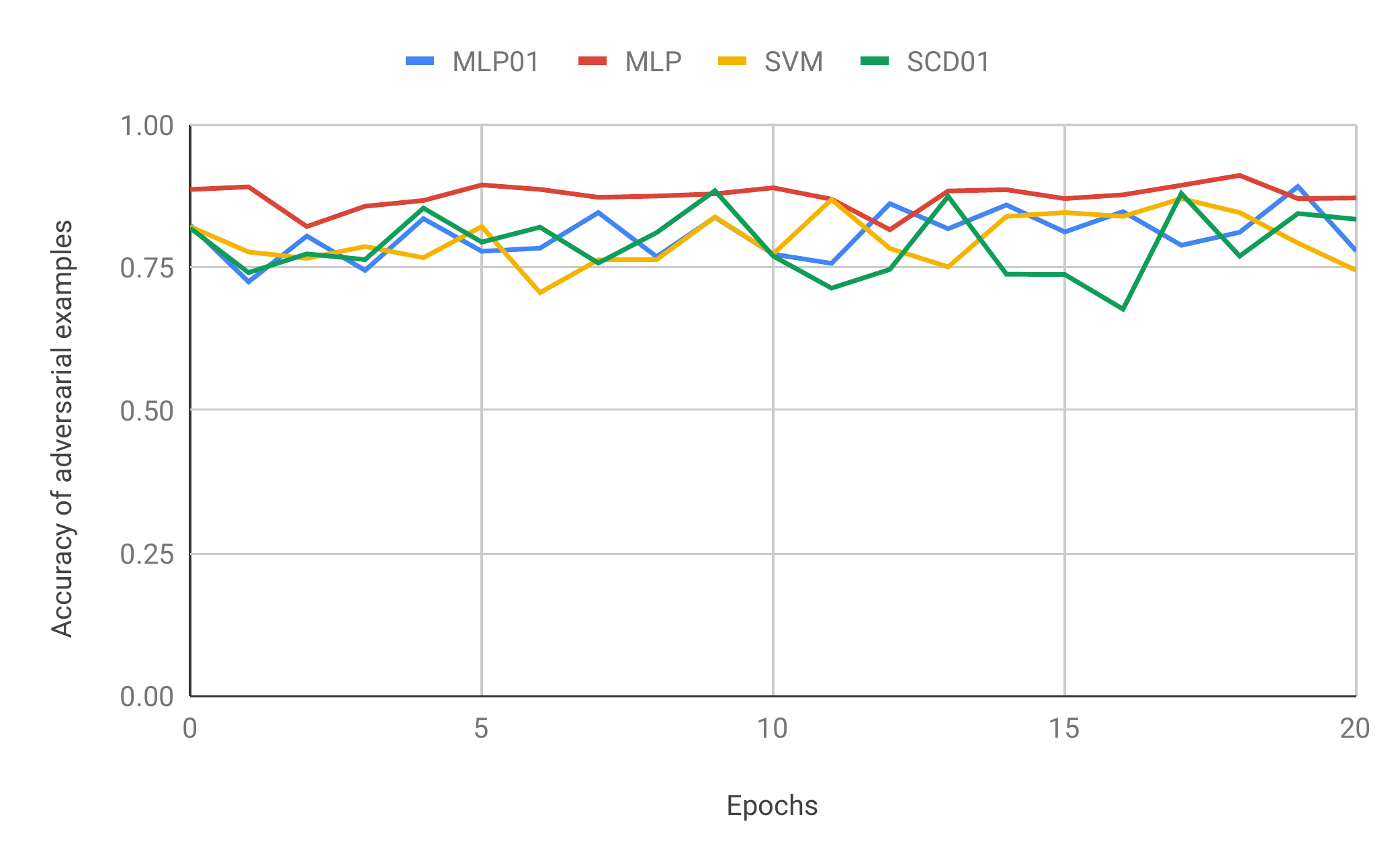}  \\
  \caption{We use SCD01 single run as the substitute model to attack single runs of the target models between only classes 0 and 1 in CIFAR10. In epoch 0 are the clean test accuracies. \label{blackbox2}}
\end{figure}

In Figure~\ref{blackbox2} we see the results of a black box attack with SCD01 single run as the substitute model attacking single runs of target models. We see that adversaries produced from this model hardly affect any of the target models in any of the epochs. Even when the target is SCD01 and trained with the same initial seed as the substitute the adversaries are ineffective. 

\begin{figure}[h!]
\begin{tabular}{c}
\includegraphics[scale=.33]{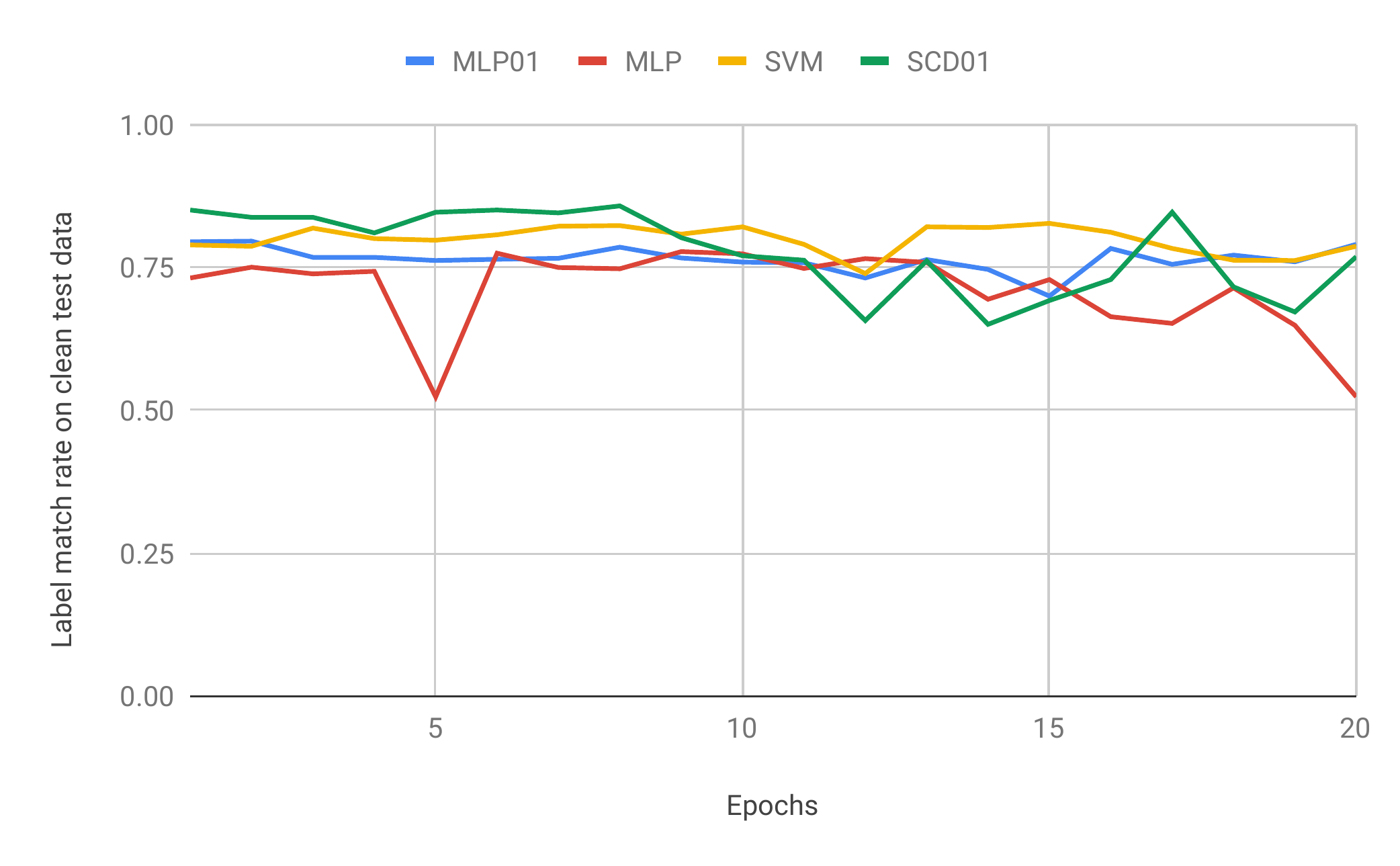}\\
\footnotesize (a) Percent of same labels between substitute and target on clean data \\
\includegraphics[scale=.33]{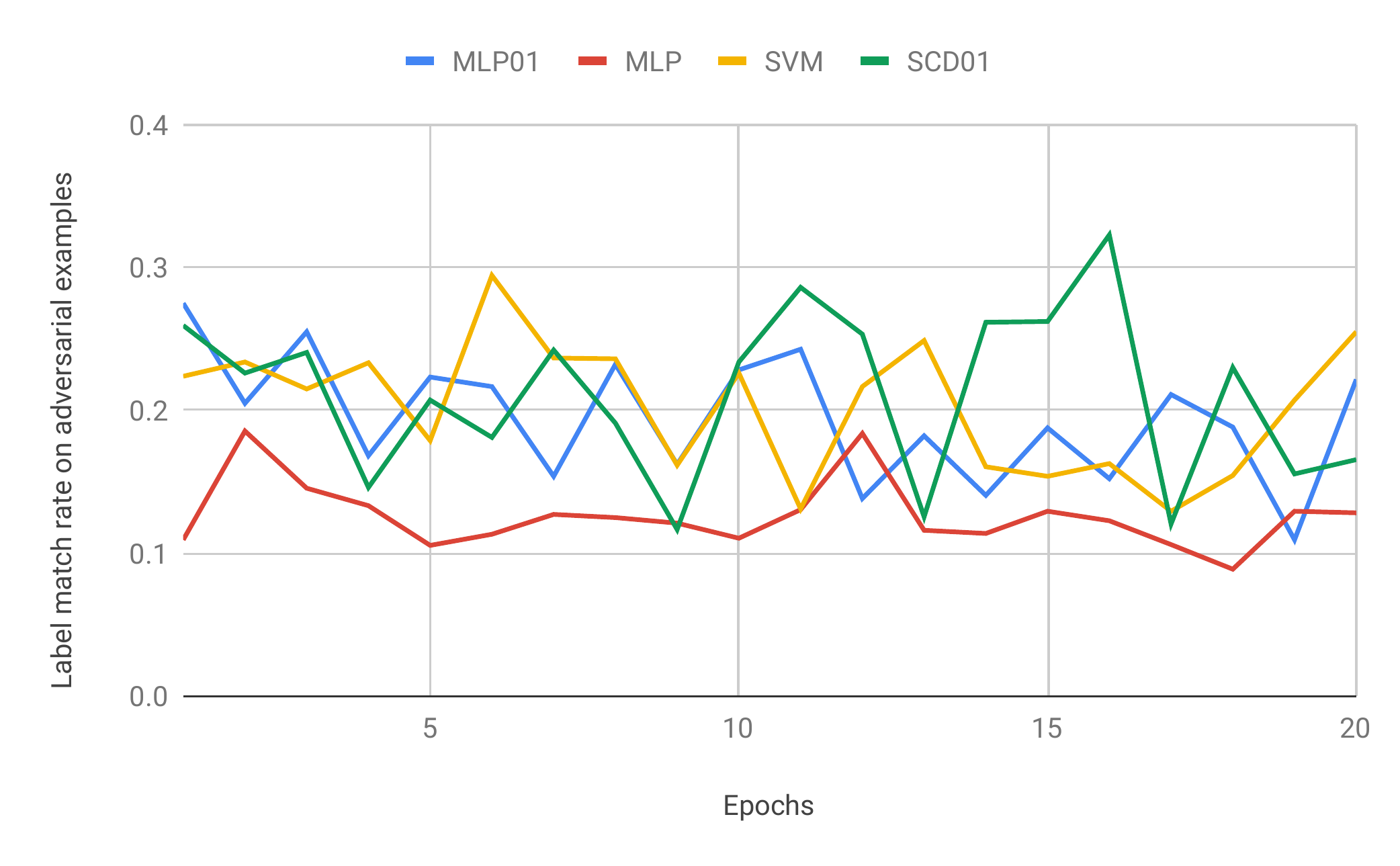}\\
\footnotesize (b) Percent of same labels between substitute and target on adversarial data \\
\end{tabular}
\caption{In (a) we see that SCD01 substitute model can approximate the target as shown in the label match rates between them. But SCD01 sourced adversaries have a lower match rate which indicates that the direction of the SCD01 boundary is very different from the targets and thus its adversaries have very little effect on the target. \label{01lossattack}}
\end{figure}

Further investigation reveals that the percentage of test data whose labels match between the 01 loss substitute and its target (known as the label match rate) is high but the label match rate on adversarial examples is much lower (shown in Figure~\ref{01lossattack}). 

Thus even though the SCD01 manages to approximate the target boundary its direction is different which gives ineffective adversaries. This is due to the non-uniqueness of 01 loss which makes single run solutions different from each other. Thus as a substitute model in black box attacks 01 loss is ineffective even in attacking itself.

\subsection{Sensitivity to training data distribution and discretized features}
The pixel distributions on MNIST shows that most pixels are near 0 and 1 whereas CIFAR10 pixel values are normally distributed \cite{weiguang2019sensitivity}. Could this be the cause of transferability between 01 loss and convex on MNIST and non-transferability between them on CIFAR10? To test this we discretize the input features by saturation \cite{weiguang2019sensitivity}: for each pixel x and fixed p we saturate it towards 1 or 0 using the formula $x^p = sign(2x-1)\frac{|2x-1|^{\frac{2}{p}}}{2} + \frac{1}{2}$. In Table~\ref{binarize} we compare white box adversary transferability between our four models on the original CIFAR10 dataset between classes 0 vs 1 and a fully binarized one with saturation parameter $p=\infty$. Even after binarizing CIFAR10 adversarial examples do not transfer effectively between 01 loss and convex models, in fact the transferability becomes harder. Thus it is likely that different boundaries caused my outliers are the cause of non-transferability on CIFAR10.

\begin{table}[!h]
  \caption{Accuracy of adversaries made by source in the first column targeting models in the top row  (class 0 vs 1). \label{binarize}}
  \centering
  \begin{tabular}{lllll} \hline
               & SVM  & SCD01 & MLP & MLP01 \\ \hline
       & \multicolumn{4}{c}{CIFAR10 original $\epsilon=.0625$} \\
   Clean &  82.2 &	81.1	& 88.7 &	84.2 \\
    SVM   &  0	& 41.3&	0.5 & 	70.1	\\
    SCD01 &   76 & 	0.8	& 86 &	84.5 \\
    MLP  &  0	& 43.5 &	0.4 & 	63.7 \\
    MLP01 &  81.7	& 80 & 	88.5 &	66.9 \\ \hline
       & \multicolumn{4}{c}{CIFAR10 fully binarized $\epsilon=.0625$} \\
   Clean &  76.6 &	79.1	& 81.2 &	78.6 \\
    SVM   &  0	&  73.9 &	6.5 & 	76.7	\\
    SCD01 &   71.1 & 	62.4	& 78.1 &	78.5 \\
    MLP  &  0.3	& 74.2 &	3 & 	76.3 \\
    MLP01 &  76.7	& 79.4 & 	81 &	73.8 \\ \hline
   \end{tabular}
\end{table}

In fact discretizing the input features increased robustness of convex and 01 loss models (as also seen previously for convex models \cite{panda2019discretization}). Below in Figure~\ref{discrete} we see the accuracy of adversarial examples when we attack models trained on the original input features and binarized ones. The clean test accuracy is slightly lower in the binarized feature space but the robustness is better (particularly for our 01 loss models). In MNIST we see 100\% robustness after binarizing input features.

\begin{figure}[!h]
  \centering
  \begin{tabular}{c}
  \includegraphics[scale=.33]{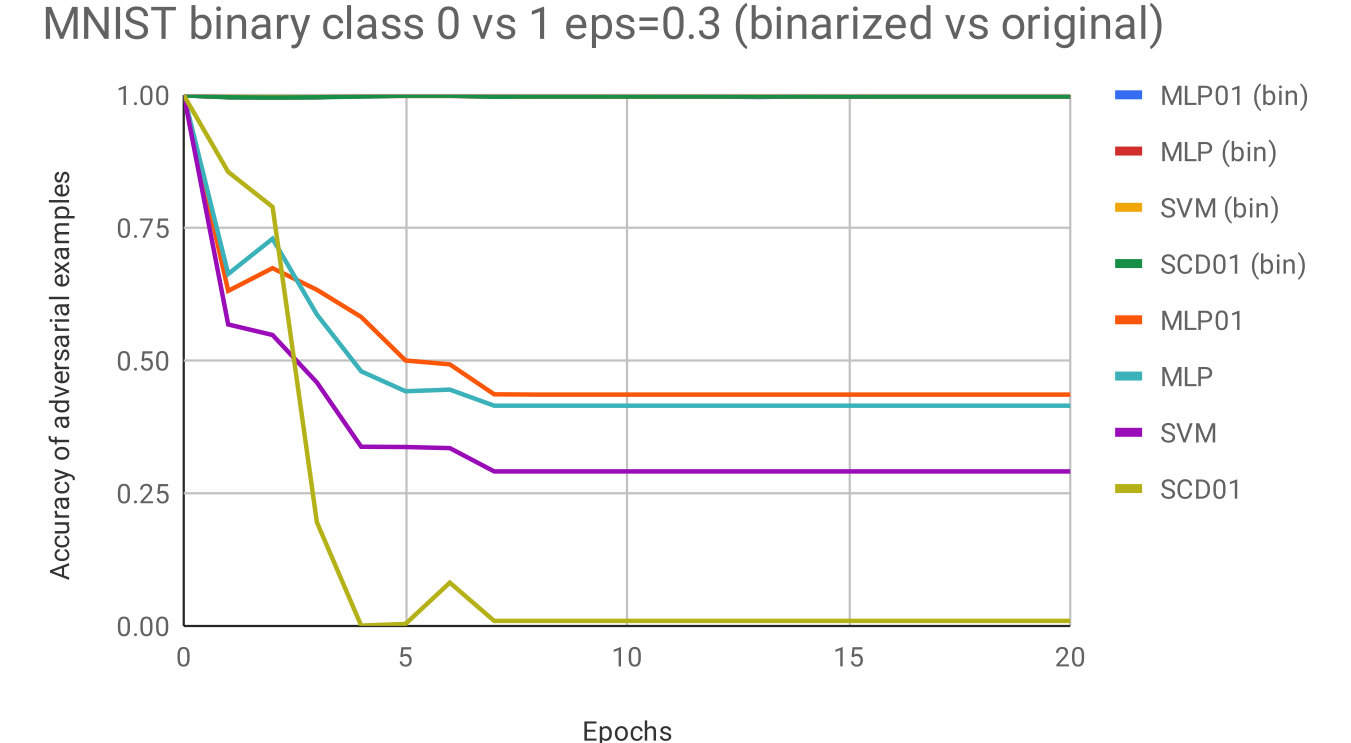} \\
  \includegraphics[scale=.33]{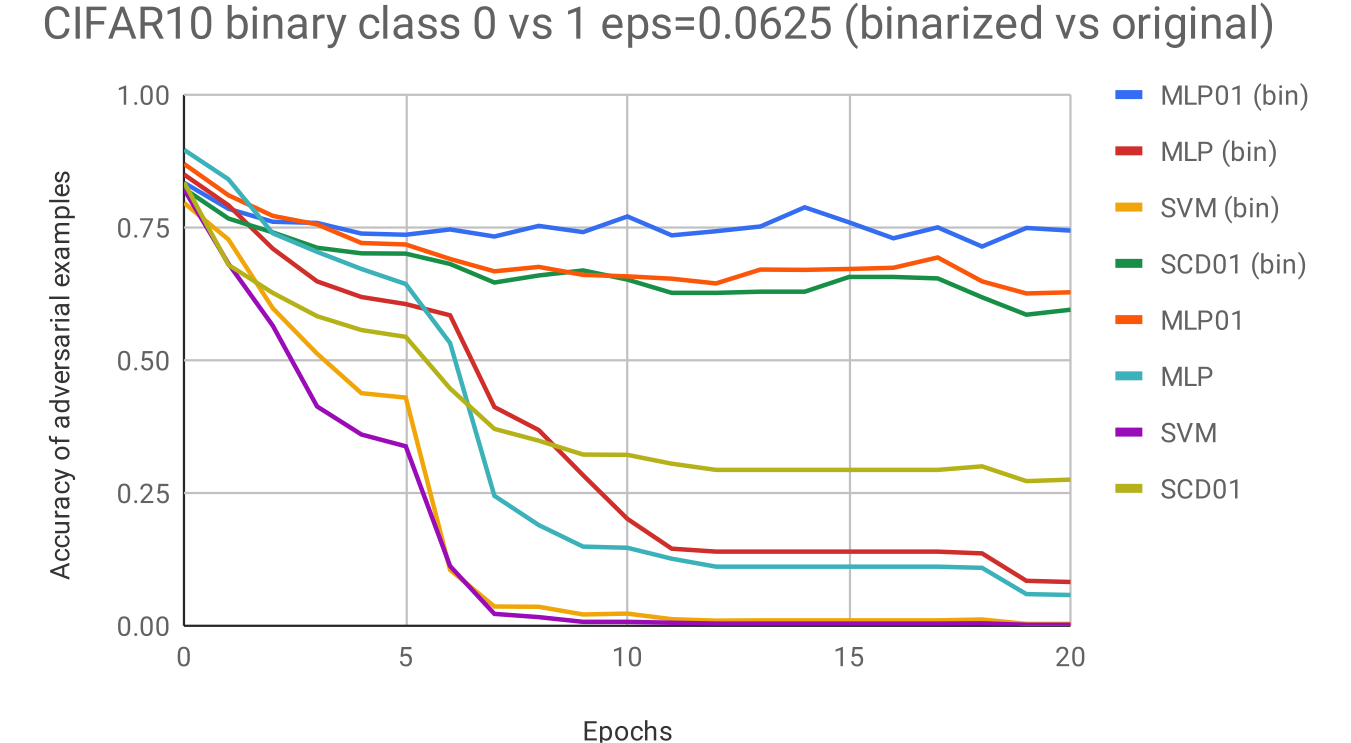} \\
 \end{tabular}
  \caption{Black box attacks with a dual 200 node hidden layer logistic loss network as the substitute model. In epoch 0 are the clean test accuracies. All models are more robust on the binarized features than the original space. \label{discrete}}
\end{figure}

\subsection{Comparison to convolutional neural networks}
We finally compare black box robustness of our 01 loss models to simple convolutional neural networks. We find the pair of CIFAR10 classes where MLP01 achieves the highest test accuracy (this turns out to be classes 6 vs 8). We then train MLP01 and MLP each with 400 hidden nodes and two convolutional neural networks. We take LeNet \cite{lecun1998gradient} and SimpleNet80 that has the convolutional layers from LeNet followed by a single hidden layer of 80 nodes. The latter is a watered down version of LeNet with fewer hidden layers.

We use a convolutional neural network with four convolutional blocks as the substitute model. In each convolutional block we have a $3\times3$ convolutional kernel followed by max pool and batch normalization. In the first, second, third, and fourth layer we have 32, 64, 128, and 256 kernels and a final layer for the output. Thus the substitute model here is much more sophisticated and powerful than the one used in our experiments above.

\begin{figure}[!h]
  \centering
  \begin{tabular}{c}
  \includegraphics[scale=.33]{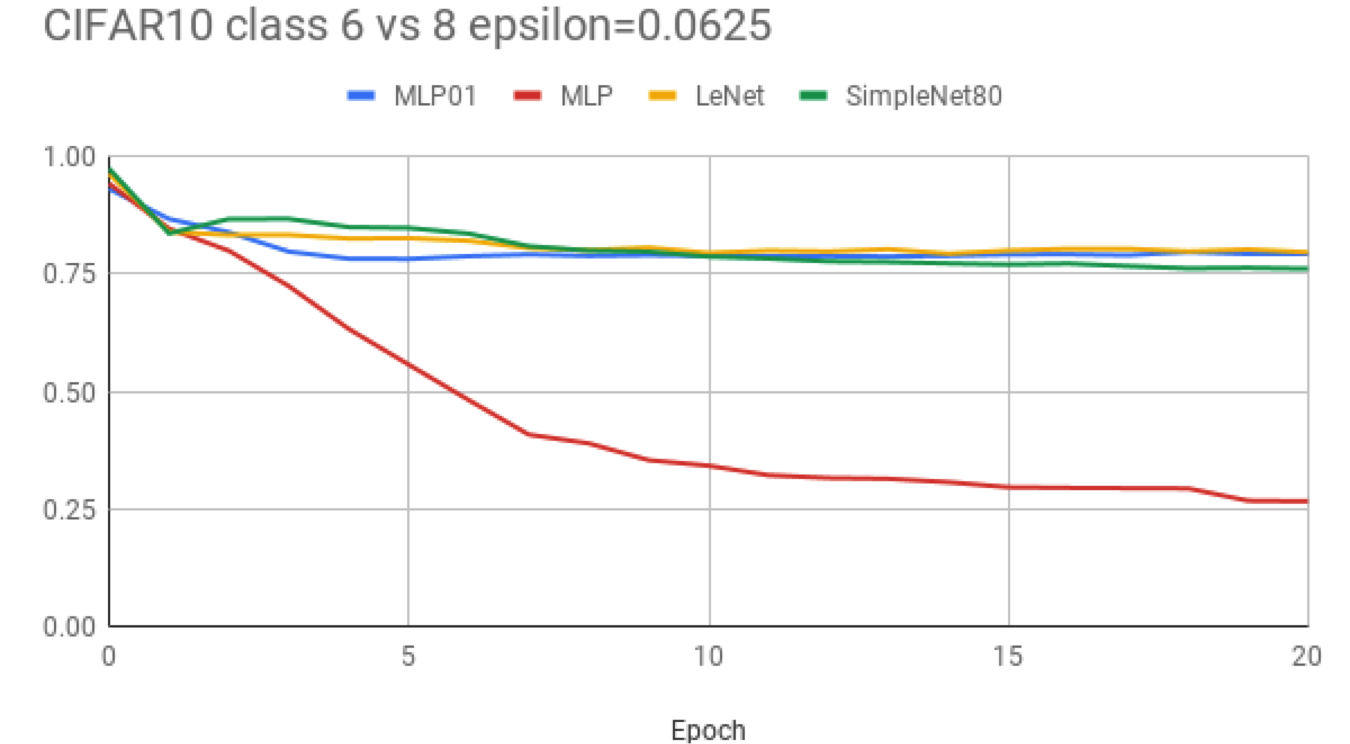} \\
  \includegraphics[scale=.33]{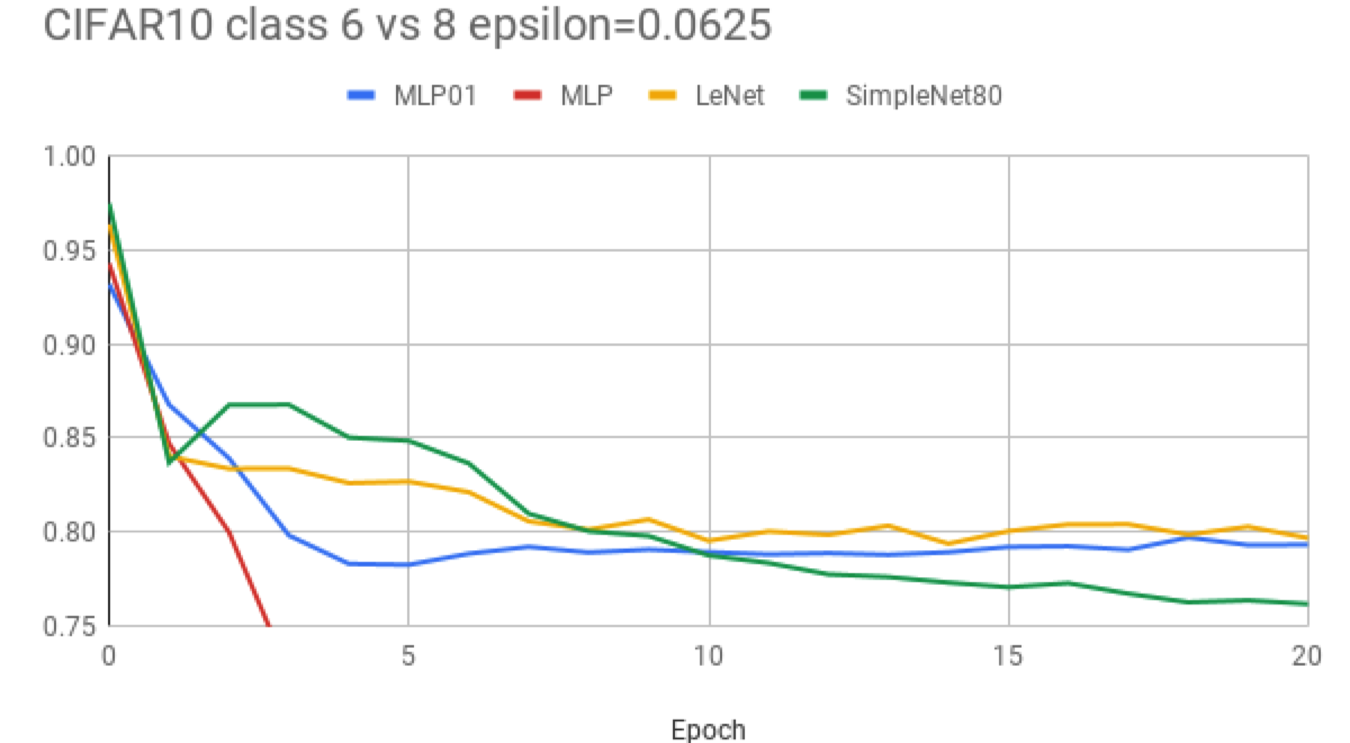} \\
 \end{tabular}
  \caption{Black box attacks with a convolutional neural network of four convolutional blocks. In epoch 0 are the clean test accuracies. We see that MLP01 is on-par with convolutional networks but MLP has much lower robustness. \label{cnn}}
\end{figure}

In Figure~\ref{cnn} we see that MLP01 has lower but comparable clean test accuracies than all the convex models including its convex counterpart MLP. However, its robustness is much higher than MLP and even surpasses that of SimpleNet80 after the tenth epoch of the substitute model training. 

\section{Discussion and Conclusion}
Binarized neural networks \cite{galloway2017attacking,courbariaux2016binarized,rastegari2016xnor} have weights and activations constrained to be near +1 and -1 (or 1 and 0) whereas our model weights are real numbers. The purpose of those networks is efficiency as opposed to robustness. Indeed we see in recent work that binarized networks offer marginal improvements in robustness to substitute model black box robustness on MNIST and none in CIFAR10 (see Tables 4 and 5 in \cite{galloway2017attacking} and Table 8 in \cite{panda2019discretization}). In our case we see large improvements over convex (non-binarized) models on CIFAR10 as well as on ImageNet. 

There is nothing to indicate that 01 loss models are robust to black box attacks that do not require substitute model training \cite{brendel2017decision,chen2019hopskipjumpattack}. These attacks are, however, computationally more expensive and require separate computations for each example. A transfer based model can be more effective (and dangerous) once it has approximated the target model boundary. 

Our work shows a lack of transferability between 01 loss and convex models on datasets like CIFAR10 and ImageNet. As a result they are more robust to substitute model black box attacks than convex models on those datasets. Interestingly the robustness is on-par with simple convolutional models that have the powerful advantage of convolutions which our 01 loss models lack. As future work 01 loss convolutions may be a promising avenue for models with high clean test accuracy and high adversarial accuracy as well.


\bibliography{my_bib}
\bibliographystyle{unsrt}

\end{document}